\documentclass{article} 
\usepackage{nips13submit_e,times}
\usepackage{hyperref}
\usepackage{url}
\usepackage{graphicx}
\usepackage{amsmath}
\usepackage{epsfig}
\usepackage{subfigure}
\usepackage{natbib}

\title{Zero-Shot Learning for Semantic Utterance Classification}

\author{
Yann N. Dauphin$^1$~~~Gokhan Tur$^2$~~~Dilek Hakkani-T\"ur$^2$~~~Larry Heck$^2$ \\
$^1$University of Montreal, Montreal, Canada\\$^2$Microsoft Research, Mountain View, CA, USA \\
}

\nipsfinalcopy 

\begin{document}

\maketitle

\begin{abstract}

We propose a novel zero-shot learning method for semantic utterance
classification (SUC). It learns a classifier $f: X \to Y$ for problems
where none of the semantic categories $Y$ are present in the training set. The
framework uncovers the link between categories and utterances through a semantic
space. We show that this semantic space can be learned by deep neural
networks trained on large amounts of search engine query log data. What's more,
we propose a novel method that can learn discriminative semantic
features without supervision. It uses the zero-shot learning framework to guide
the learning of the semantic features. We demonstrate the effectiveness of the
zero-shot semantic learning algorithm on the SUC dataset collected by 
\citep{DCN}.  Furthermore, we achieve state-of-the-art results by combining the
semantic features with a supervised method.

\end{abstract}

\section{Introduction}

Conversational understanding systems aim to automatically classify user requests
into predefined semantic categories and extract related
parameters~\citep{SLUBook}. For instance, such a system might classify the
natural language query \textit{``I want to fly from San Francisco to
New York next Sunday"} into the semantic domain \textit{flights}. This is known
as semantic utterance classification (SUC). Typically, these systems use supervised
classification methods such as Boosting~\citep{boostexter}, support vector machines
(SVMs)~\citep{patrick-gokhan}, or maximum entropy
models~\citep{Sibel-IEEE-TASLP-2008}. These methods can produce
state-of-the-art results but they require significant amounts of labelled data.
This data is mostly obtained through manual labor and becomes costly as the
number of semantic domains increases. This limits the applicability of these
methods to problems with relatively few semantic categories.

We consider two problems here. First, we examine the problem of predicting the semantic domain
of utterances without having seen examples of any of the domains. Formally, the
goal is to learn a classifier $f: X \to Y$ without any values of $Y$ in the
training set. In constrast to traditional SUC systems, adding a domain is as
easy as including it in the set of domains.
This is a form of zero-shot learning \citep{palatucci2009zero} and is possible 
through the use of a knowledge base of semantic properties of the classes to extrapolate
to unseen classes. Typically this requires seeing examples of at least some of
the semantic categories. Second, we consider the problem of easing the task
of supervised classifiers when there are only few examples per domain. This is
done by augmenting the input with a feature vector $H$ for a classifier
$f: (X, H) \to Y$. The difficulty is that $H$ must be learned without any
knowledge of the semantic domains $Y$.

In this paper, we introduce a zero-shot learning framework for SUC where none of
the classes have been seen. We propose to use a knowledge base which can output
the semantic properties of both the input and the classes. The classifier
matches the input to the class with the best matching semantic features.
We show that a knowledge-base of semantic properties can be learned
automatically for SUC by deep neural networks using large amounts of data. The recent
advances in deep learning have shown that deep networks trained at large scale
can reach state-of-the-art results. We use the Bing search query click logs,
which consists of user queries and associated clicked URLs. We hypothesize that
the clicked URLs reflect high level meaning or intent of the queries.
Surprinsingly, we show that is is possible to learn semantic properties which
are discriminative of our unseen classes without any labels. We call this method
zero-shot discriminative embedding (ZDE). It uses the zero-shot learning
framework to provide weak supervision
during learning. Our experiments show that the zero-shot learning framework for
SUC yields competitive results on the tasks considered. We demonstrate that
zero-shot discriminative embedding produces more discriminative semantic properties.
Notably, we reach state-of-the-art results by feeding these features to an SVM.

In the next section, we formally define the task of semantic utterance
classification. We provide a quick overview of zero-shot learning in
Section~\ref{sec:zeroshot}. Sections~\ref{sec:zs} and \ref{sec:dnn} present
the zero-shot learning framework and a method for learning semantic features
using deep networks. Section~\ref{sec:clu} introduces the zero-shot
discriminative embedding method. We review the related
work on this task in Section~\ref{sec:rel} In
Section~\ref{sec:exp} we provide experimental results.

\section{Semantic Utterance Classification}

The semantic utterance classification (SUC) task aims at classifying a
given speech utterance $X_r$ into one of $M$ semantic classes,
$\hat{C}_r\in\mathcal{C}=\{C_1,\ldots,C_M\}$ (where $r$ is the utterance
index). Upon the observation of $X_r$, $\hat{C}_r$ is chosen so that
the class-posterior probability given $X_r$, $P(C_r|X_r)$, is
maximized. More formally, $\hat{C}_r=\hbox{arg}\max_{C_r}P(C_r|X_r)$.

Semantic classifiers need to allow significant utterance variations. A
user may say \textit{``I want to fly from San Francisco to New York
next Sunday"} and another user may express the same information by
saying \textit{``Show me weekend flights between JFK and SFO"}. 
Not only is there no {\it a priori} constraint on what
the user can say, these systems also need to generalize well from a
tractably small amount of training data. On
the other hand, the command \textit{``Show me the weekend snow forecast"}
should be interpreted as an instance of another semantic class, say,
``{\em Weather}." In order to do this, the selection of the feature functions
$f_i(C,W)$ aims at capturing the relation between the class $C$ and
word sequence $W$. Typically, binary or weighted $n$-gram features,
with $n=1,2,3$, to capture the likelihood of the $n$-grams, are
generated to express the user intent for the semantic class
$C$~\citep{SLUBook-chapter}. 
Once the features are extracted from the text, the task becomes a
text classification problem.  Traditional text categorization
techniques devise learning methods to maximize the probability of
$C_r$, given the text $W_r$; i.e., the class-posterior
probability $P(C_r|W_r)$. 

\section{Zero-shot learning}
\label{sec:zeroshot}

In general, zero-shot learning \citep{palatucci2009zero} is concerned with
learning a classifier $f: X \to Y$ that can predict novel values of $Y$ not
present in the training set. It is an important problem setting for tasks where
the set of classes is large and in cases where the cost of labelled examples is
high. It has found application in vision where the number of classes can be very
large \citep{41473}.

A zero-shot learner uses semantic knowledge to extrapolate to novel classes.
Instead of predicting the classes directly, the learner predicts semantic
properties or features of the input. Thanks to a knowledge-base of semantic
features for the classes it can match the inputs to the classes.

The semantic feature space is a euclidean space of $d$ dimensions. Each
dimension encodes a semantic property. In vision for instance, one dimension
might encode the size of the object, another the color. The knowledge base
$\mathcal{K}$ stores a semantic feature vector $H$ for each of the classes.
The zero-shot classifier $f = m\circ n$ is the composition of two classifiers.
The first classifier $m: X \to H$ predicts the semantic properties of the input.
The training set is found by replacing the class values in the training set by
their semantic features. The second classifier $n: H \to Y$ matches the semantic
code to the class. This can be done by a $k$-NN classifier.

In applying zero-shot learning to semantic utterance classification there are
several challenges. The framework described by \citep{palatucci2009zero}
requires some of the classes to be present in the training data in order
to train the $m$ classifier. We are interested in the setting where none of
classes have training data. Furthermore, an adequate knowledge-base must be
found for SUC.

\section{Zero-Shot Learning for Semantic Utterance Classification}\label{sec:zsl}
\label{sec:zs}

In this section, we introduce a zero-shot learning framework for SUC where none
of the classes are seen during training. It is based on the observation that in
SUC both the semantic categories and the inputs reside in the same semantic
space. In this framework, classification can be done by finding the best
matching semantic category for a given input.

Semantic utterance classification is concerned with finding the semantic
category for a natural language utterance. Traditionally, conversational systems
learn this task using labelled data. This overlooks the fact that
classification would be much easier in a space that reveals the semantic
meaning of utterances. Interestingly, the semantics of
language can be discovered without labelled data. What's more,
the name of semantic classes are not chosen randomly. They are in the same
language as the sentences and are often chosen
because they describe the essence of the class. These two facts can easily be used
by humans to classify without task-specific labels. For instance, it is
easy to see that the utterance \emph{the accelerator has exploded} belongs more to
the class \emph{physics} than \emph{outdoors}. This is the very human ability
that we wish to replicate here.

\begin{figure}[h]
  \centering
    \includegraphics[height=6cm]{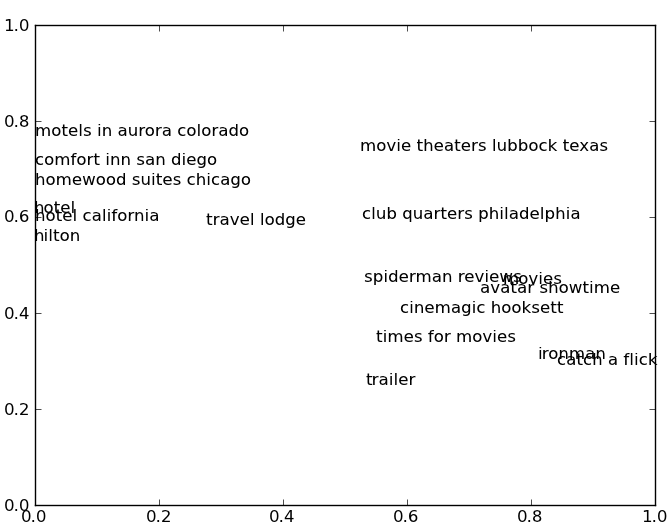}
  \caption{\label{fig:embedding} Visualization of the 2d semantic space learned by
  a deep neural net. We see that the two axis differentiate between phrases
  relating to hotels and movies. More details in Section \ref{sec:exp}.}
\end{figure}

We propose a framework called zero-shot semantic learning (ZSL) that leverages
these observations. In this framework, the knowledge-base $\mathcal{K}$ is a
function which can output the semantic properties of any sentence. The
classification procedure can be done in one step because both the input and the
categories reside in the same space. The zero-shot classifier finds the 
category which best matches the input. More formally, the zero-shot classifier
is given by
\begin{eqnarray}\label{eqn:zsl}
P(C_r|X_r) = \frac{1}{Z} e^{- | \mathcal{K}(X_r) - \mathcal{K}(C_r) |}
\end{eqnarray}
where $Z = \sum_{C} e^{- | \mathcal{K}(X_r) - \mathcal{K}(C)|}$ and $|x-y|$ is
a distance measure like the euclidean distance. The knowledge-base maps
the input $\mathcal{K}(X_r)$ and the category $\mathcal{K}(X_r)$ in a space
that reveals their meaning. An example 2d semantic space is given in Figure 
\ref{fig:embedding} which maps sentences relating to movies close to each
other and those relating to hotels further away. In this space, given the
categories \emph{hotel} and \emph{movies}, the sentence
\emph{motels in aurora colorado} will be classified to \emph{hotel} because
$\mathcal{K}(\textit{motels in aurora colorado})$ is closer to
$\mathcal{K}(\textit{hotel})$.

This framework will classify properly if
\begin{itemize}
    \item The semantics of the language are properly captured by $\mathcal{K}$.
    In other words, utterances are clustered according to their meaning.
    \item The class name $C_r$ describes the semantic core of the class well.
    Meaning that $\mathcal{K}(C_r)$ resides close to the semantic representation
    of sentences of that class.
\end{itemize}

The success of this framework rests on the quality of the knowledge-base
$\mathcal{K}$. Following the success of learning methods with language, we
are interested in learning this knowledge-base from data. Unsupervised
learning methods like LSA, and LDA have had some success but it is hard
to ensure that the semantic properties will be useful for SUC.

\section{Learning Semantic Features for SUC using Deep Nets}
\label{sec:dnn}

In this section, we describe a method for learning a semantic features for SUC
using deep networks trained on Bing search query click logs. We use the query
click logs to define a task that makes the networks learn the meaning or intent
behind the queries. The semantic features are found at the last hidden layer
of the deep neural network.

Query Click Logs (QCL) are logs of unstructured text including both the users
queries sent to a search engine and the links that the users clicked on from the
list of sites returned by that search engine. Some of the challenges in
extracting useful information from QCL is that the feature space is very high
dimensional (there are thousands of url clicks linked to many queries), and
there are millions of queries logged daily.

We make the mild hypothesis that the website clicked following a query reveals
the meaning or intent behind a query. The queries which have
similar meaning or intent will map to the same website. For
example, it is easy to see that queries associated with
the website \emph{imdb.com} share a semantic connection to movies.

\begin{figure}[h]
    \centerline{\includegraphics[height=4cm,width=8cm]{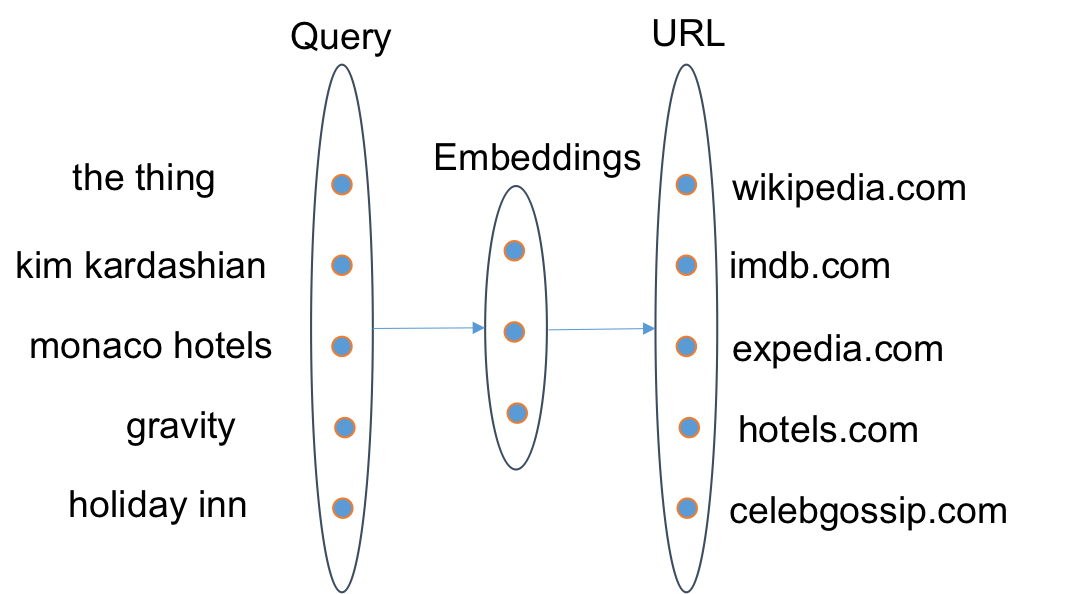}}
    \caption{Depiction of the deep network from queries to URLs. }
  \label{fig:qcl}
\end{figure}

We train the network with the query as input and the website
as the output (see Figure \ref{fig:qcl}). This learning scheme is
inspired by the neural language models
\citep{Bengio-scholarpedia-2007} who learn word embeddings
by learning to predict the next word in a sentence. The idea is that the last
hidden layer of the network has to learn an embedding space which is
helpful to classification. To do this, it will map similar
inputs in terms of the classification task close in the embedding space. The key
difference with word embeddings methods like \citep{Bengio-scholarpedia-2007} is that we are learning sentence-level embeddings.

We train deep neural networks with softmax output units and
rectified linear hidden units. The inputs $X_r$ are
queries represented in bag-of-words format. The labels
$Y_r$ are the index of the website that was clicked. We
train the network to minimize the negative log-likelihood
of the data $\mathcal{L}(X, Y) = -\log P(Y_r|X_r)$.

The network has the form
\[
P(Y=i|X_r) = \frac{e^{W^{n+1}_i H^{n}(X_r) + b^{n+1}_i}}{\sum_j e^{W^{n+1}_j H^{n}(X_r) + b^{n+1}_j}}
\]
The latent representation function $H^n$ is composed on $n$ hidden layers
\begin{eqnarray}
H^{n}(X_r) & = & \max(0, W^n H^{n-1}(X_r) + b^n) \nonumber \\
H^{1}(X_r) & = & \max(0, W^1 X_r + b^1) \nonumber
\end{eqnarray}
We have a set of weight matrices $W$ and biases $b$ for each layer
giving us the parameters $\theta = \{ W^1, b^1, \dots, W^{n+1}, b^{n+1} \}$
for the full network. We train the network using
stochastic gradient descent with minibatches.

The knowledge-base function is given by the last hidden layer
$\mathcal{K} = H^n(X_r)$. In this scheme, the embeddings are used as the
semantic properties of the knowledge-base. However, it is not clear that the
semantic space will be discriminative of the semantic categories we care about
for SUC.

\section{Learning Discriminative Semantic Features without Supervision}
\label{sec:clu}

We introduce a novel regularization that encourages deep networks to learn
discriminative semantic features for the SUC task without labelled data. More
precisely, we define a clustering measure for the semantic classes using the
zero-shot learning framework of Section \ref{sec:zsl}. We hypothesize the 
classes are well clustered hence we minimize this measure.

In the past section, we have described a method for learning semantic features
using query click logs. The features are given by finding the best semantic
space for the query click logs task. In general, there might be a mismatch
between what qualifies as a good semantic space for the QCL and SUC tasks. For
example, the network might learn an embedding that clusters sentences of the
category \emph{movies} and \emph{events} close together because they both relate to
activities. In this case the features would have been more discriminative if the
sentences were far from each other. However, there is no pressure for the
network to do that because it doesn't know about the SUC task.

\begin{figure*}[h]
    \vspace{-3mm}
    \centering
    \subfigure{\includegraphics[width=0.49\linewidth]{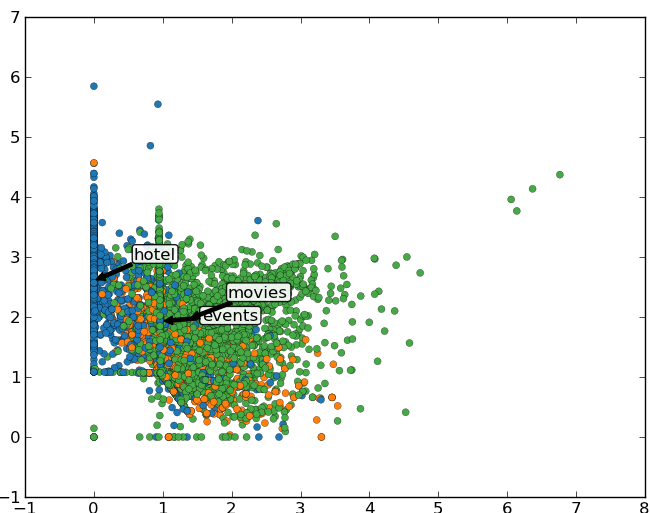}}
    \subfigure{\includegraphics[width=0.49\linewidth]{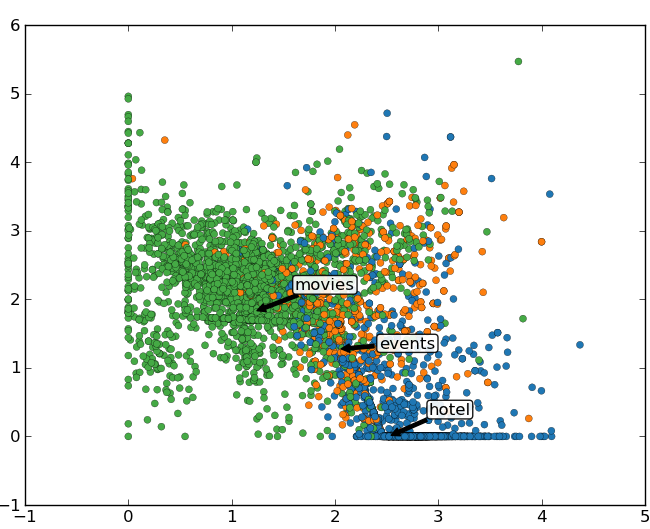}}
    \caption{\label{fig:clustering} Visualization of an actual 2d embedding space
    learned by a DNN (left) and DNN trained with ZDE (right). The points are
    sentences with different colors for each class and the arrows point to the
    location of the class name in the embedding space. ZDE significantly
    improves the clustering of the classes. More details in Section \ref{sec:exp}.}
\end{figure*}

This problem could have been addressed by multi-task or semi-supervised learning
methods if we had access to labelled data. Research has shown adding even a
little bit of supervision is often helpful \citep{Larochelle-jmlr-2009}. The
simplest solution would be to train the network on the QCL and SUC task
simultaneously. In other words, we would train the network to minimize the
sum of the QCL objective $-\log P(Y|X)$ and the SUC objective $-\log P(C|X)$. 
This would allow the model to leverage the large amount of QCL data while learning a
better representation for SUC. We cannot miminize $-\log P(C|X)$ but we can
minimize a similar measure which does not require labels.

We can measure the overlap of the semantic categories using the conditional
entropy
\begin{eqnarray}\label{eqn:entropy}
H(P(C_r|X_r)) & = & E[I(P(C_r|X_r))] \\
& = & E[-\sum_i P(C_r=i|X_r)\log P(C_r=i|X_r)]. \nonumber
\end{eqnarray}
The measure is lowest when the overlap is small. Interestingly, calculating the
entropy does not require labelled data. We can recover a zero-shot classifier
$P(C|X)$ from the semantic space using Equation \ref{eqn:zsl}. The entropy
$H(P(C_r|X_r))$ of this classifier measures the clustering of the categories
in the semantic space. Spaces with the lowest entropy are those where the
examples $\mathcal{K}(X_r)$ cluster around category names $\mathcal{K}(C_r)$ and
where the categories have low-overlap in the semantic space. Figure
\ref{fig:clustering} illustrates a semantic space with high conditional entropy
on the left, and one with a low entropy on the right side.

Zero-shot Discriminative Embedding (ZDE) combines the embedding method of
Section \ref{sec:dnn} with the minimization of the entropy of a zero-shot
classifier on that embedding. The objective has the form
\begin{eqnarray}\label{eqn:spc}
\mathcal{L}(X, Y) = -\log P(Y|X) + \lambda H(P(C|X)).
\end{eqnarray}
The variable $X$ is the input, $Y$ is the website that was clicked, $C$ is a
semantic class. The hyper-parameter $\lambda$ controls the strength of entropy
objective in the overall objective. We find this value by cross-validation.

\section{Related work}
\label{sec:rel}

Early work on spoken utterance classification has been done mostly for
call routing or intent determination system, such as the AT\&T How May
I Help You? (HMIHY) system~\citep{hmihy}, relying on salience phrases,
or the Lucent Bell Labs vector space model~\citep{lucenthmihy}.  Typically word
$n$-grams are used as features after preprocessing with generic
entities, such as dates, locations, or phone numbers. Because of the
very large dimensions of the input space, large margin classifiers
such as SVMs~\citep{patrick-gokhan} or Boosting~\citep{boostexter} were
found to be very good candidates. Deep learning methods have first been used
for semantic utterance classification by Sarikaya et al.~\citep{ruhi-dbn:2011}.
Deep Convex Networks (DCNs)~\citep{DCN} and Kernel DCNs
(K-DCNs)~\citep{deng-SLT12} have also been applied to SUC.  K-DCNs allow the use of
kernel functions during training, combining the power of kernel based
methods and deep learning.  While both approaches resulted in
performances better than a Boosting-based baseline, K-DCNs have shown
significantly bigger performance gains due to the use of query click
features.

Entropy minimization \citep{NIPS2005_519} is a semi-supervised learning framework
which also uses the conditional entropy. In this framework, both labelled and
unlabelled data are available, which is an important difference with ZDE. In \citep{NIPS2005_519}, a
classifier is trained to minimize its conditional likelihood and its conditional
entropy. ZDE avoids the need for labels by minimizing the entropy of a zero-shot
classifier. \citep{NIPS2005_519} shows that this approach produces good results
especially when generative models are mispecified.

\section{Experiments}
\label{sec:exp}

In this section, we evaluate the zero-shot semantic learning framework and
the zero-shot discriminative embedding method proposed in the previous sections.

\subsection{Setup}

We have gathered a month of query click log data from Bing to learn the
embeddings. We restricted the websites to the the 1000 most
popular websites in this log. The words in the bag-of-words vocabulary
are the 9521 found in the supervised SUC task we will use. All queries
containing only unknown words were filtered out. We found that using a
list of stop-words improved the results. After these restrictions, the
dataset comprises 620,474 different queries.

We evaluate the performance of the methods for SUC on the dataset
gathered by \citep{DCN}. It was compiled from utterances by users
of a spoken dialog system. There are 16,000 training utterances,
2000 utterances for validation and 2000 utterances for testing.
Each utterance is labelled with one of 25 domains.

The hyper-parameters of the models are tuned on the validation set.
The learning rate parameter of gradient descent is found by grid
search with $\{ 0.1, 0.01, 0.001 \}$. The number of layers is
between 1 and 3. The number of hidden units is kept constant
through layers and is found by sampling a random number from 300 to
800 units. We found that it was helpful to regularize the networks
using dropout \citep{Hinton-et-al-arxiv2012}. We sample the dropout
rate randomly between 0\% dropout and 20\%. The $\lambda$ of the
zero-shot embedding method is found through grid-search with
$\{ 0.1, 0.01, 0.001 \}$. The models are trained on a cluster
of computers with double quad-core Intel(R) Xeon(R) CPUs with
2.33GHz and 8Gb of RAM. Training either the ZDE method
on the QCL data requires 4 hours of computation time.

\subsection{Results}

First, we want to see what is learned by the embedding method
described in Section \ref{sec:dnn}. A first step is to look at the nearest
neighbor of words in the embedding space. Table \ref{tbl:words} shows the
nearest neighbours of specific words in the embedding space. We
observe that the neighbors of the words al share the semantic domain of the
word. This confirms that the network learns some semantics of the language.

\begin{table} [h]
\centerline{
\begin{tabular}{|c|c|c|c|c|c|}
\hline
Restaurant & Hotel & Flight & Events & Transportation \\
\hline  \hline
steakhouse & suites & airline & festivals & distributing \\
diner & hyatt & airfaire & upcoming & dfw \\
seafood & resorts & plane & fireworks & petroleum \\
tavern & ramada & baggage & happening & hospitality \\
\hline
\end{tabular}}
\caption{\label{tbl:words} {\it Nearest neighbours in the embedding space. Each column displays
the 5 nearest neighbours of the word at the top. We can see that the embedding captures
the semantics of the words.}}
\end{table}

We can better visualize the embedding space using a network with a special
architecture. Following \citep{Hinton+Salakhutdinov-2006}, we train deep networks
where the last hidden layer contains only 2 dimensions. The depth allows the
network to progressively reduce the dimensionality of the data. This approach
enables us to visualize exactly what the network has learned. Figure 
\ref{fig:embedding} shows the embedding a deep network with 3 layers (with
size 200-10-2) trained on the QCL task. We observe that the embedding
distinguishes between sentences related to movies and hotels. In Figure
\ref{fig:clustering}, we compare the embedding spaces of a DNN trained on the
QCL (left) and a DNN trained using ZDE (right) both with hidden layers of sizes
200-10-2. The comparison suggests that minimizing the conditional entropy of the
zero-shot classifier successfully improves the clustering.

\begin{table*} [h]
\centerline{
\begin{tabular}{|l|c|c|c|c|c|}
\hline
Method & Restaurant & Hotel & Flight & Events & Transportation \\
\hline  \hline
ZSL with Bag-of-words & 0.616 & 0.641 & 0.683 & 0.559 & 0.5 \\
ZSL with $p(Y|X)$ (LR) & 0.779 & 0.821 & 0.457 & 0.677 & 0.472 \\
ZSL with $p(Y|X)$ (DNN) & 0.838 & 0.862 & 0.46 & 0.631 & 0.503 \\
ZSL with DNN Embedding & 0.858 & 0.935 & 0.870 & 0.727 & 0.667 \\
ZSL with ZDE Embedding & {\bf 0.863} & {\bf 0.940} & {\bf 0.906} & {\bf 0.841} & {\bf 0.826} \\
Representative URL heuristic (DNN) & 0.798 & 0.892 & 0.769 & 0.707 & 0.577\\
\hline
\end{tabular}}
\caption{\label{tbl:auc} {\it Comparison of several zero-shot semantic learning methods for 5
semantic classes. Our proposed
zero-shot learning system with DNN embeddings outperforms other approaches.}}

\end{table*}

Second, we want to confirm that good classification results can be achieved
using zero-shot semantic learning. To do this, we evaluate the classification
results of our method on the SUC task. Our results are
given in Table \ref{tbl:auc}. The performance is measured
using the AUC (Area under the curve of the precision-recall curve) for which
higher is better. We compare our ZDE method against various means of obtaining
the semantic features $H$. We compare with using the bag-of-words representation
(denoted \emph{ZSL with Bag-of-words}) as semantic features.
\emph{ZSL with $p(Y|X)$ (LR)} and \emph{ZSL with $p(Y|X)$ (DNN)} are models
trained from the QCL to predict the website associated with queries. The 
semantic features are the vector of probability that each website is associated
with the query. \emph{ZSL with $p(Y|X)$ (LR)} is a logistic regression model,
\emph{ZSL with $p(Y|X)$ (DNN)} is a DNN model. We also compare
with a sensible heuristic method denoted \emph{Representative URL heuristic}.
For this heuristic, we associate each semantic category with a representative website
(i.e. \emph{flights} with \emph{expedia.com}, \emph{movies} with {imdb.com}).
We train a DNN using the QCL to predict which of these websites is clicked given an utterance.
The semantic category distribution $P(C|X)$ is the probability
that each associated website was clicked. Table \ref{tbl:auc} shows that
the proposed zero-shot learning method
with ZDE achieves the best results. In particular, ZDE improves performance by a
wide margin for hard categories like \emph{transportation}. These results
confirm the hypothesis behind both ZSL and the ZDE method.

\begin{figure}[t]
  \centering
    \includegraphics[height=7cm]{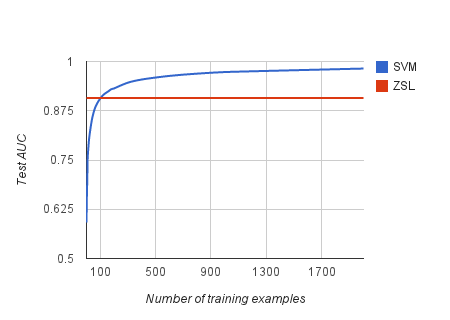}
  \caption{\label{fig:svm} Comparison between the proposed zero-shot learning
  method and an SVM trained with increasing amount of examples. The curve shows
  that ZSL compares favorably with SVMs when there are few labels.}
\end{figure}

We also compare the zero-shot learning system with a supervised SUC system. We
compare ZSL with a linear SVM. The task is identify utterances of the
\emph{restaurant} semantic class. Figure \ref{fig:svm} shows the performance of
the linear SVM as the number of labelled training examples increases. The performance of
ZSL is shown as a straight line because it does not use labelled data. Predictably, the SVM
achieves better results when the labelled training set is large. However, ZSL
achieves better performance in the low-data regime. This confirms that ZSL
can be useful in cases where labelled data is costly, or the number of classes
is large.

\begin{table} [h]
\centerline{
\begin{tabular}{|c|c|c|c|c|}
\hline
Features & Kernel DCN & SVM \\
\hline  \hline
Bag-of-words & 9.52\% & 10.09\% \\
QCL features \citep{dilek:2011}  & {\bf 5.94\%} & 6.36\% \\
DNN urls &  & 6.88\% \\
DNN embeddings  &  & 6.2\% \\
ZDE embeddings &  & {\bf 5.73\%} \\
\hline
\end{tabular}}
\caption{\label{tbl:zsl} {\it Test error rate of various methods on the SUC
task. The best results are achieved with by augmenting the input with ZDE
embeddings.}}
\end{table}

Finally, we consider the problem of using semantic features $H$ to increase
the performance of a classifier $f: (X, H) \to Y$. The input X is a bag-of-words
representation of the utterances. We compare with state-of-the-art approaches
in Table \ref{tbl:zsl}. The state-of-the-art method is the Kernel DCN on QCL
features with 5.94\% test error. However, we train using the more scalable
linear SVM which leads to 6.36\% with the same input features. The linear SVM
is better to compare features because it cannot non-linearly transform the
input by itself. Using the embeddings learned from the QCL data as described
in Section \ref{sec:zsl} yields 6.2\% errors. Using zero-shot discriminative
embedding further reduces the error t 5.73\%.

\section{Conclusion}

We have introduced a zero-shot learning framework for SUC. The proposed method
learns a knowledge-base using deep networks trained on large amounts of search
engine query log data. We have proposed a novel way to learn embeddings that
are discriminative without access to labelled data. Finally, we have shown
experimentally that these methods are effective.

\bibliography{all,aigaion,ml}
\bibliographystyle{natbib}

\end{document}